\pdfoutput=1

\documentclass[11pt]{article}

\usepackage[]{acl}

\usepackage{times}
\usepackage{latexsym}

\usepackage[T1]{fontenc}

\usepackage[utf8]{inputenc}

\usepackage{microtype}

\usepackage{inconsolata}

\usepackage{graphicx}
\usepackage{amsmath}
\usepackage{cleveref}
\usepackage{tcolorbox}
\usepackage{xcolor,colortbl}
\usepackage{caption}
\usepackage{booktabs,subcaption,amsfonts,dcolumn}
\usepackage{pifont}
\usepackage{ulem} 
\usepackage{tikz}
\usepackage{bbding}
\usepackage{cancel}
\usepackage{graphicx}
\usepackage{hyperref}  
\usepackage{enumitem}
\usepackage{algorithm}
\usepackage{algpseudocode}
\usepackage{multirow}
\usepackage{booktabs}
\usepackage{enumitem}
\usepackage{caption,subcaption}
\usepackage{graphicx}
\usepackage{xcolor, soul, colortbl}
\usepackage{twemojis}

\newtcolorbox{mybox}[1]{colback=gray!10!white,colframe=gray!75!black, boxrule=0.1mm,fonttitle=\bfseries,title=#1}
\newcommand{\grayc}[0]{\cellcolor[rgb]{0.957,0.957,0.957}}

\newcommand{\dataname}{\textit{\textbf{CliniDial}}}

\definecolor{greenLight}{RGB}{231, 243, 231}  
\definecolor{greenDark}{RGB}{204, 229, 204}   

\definecolor{lb}{RGB}{240, 245, 255}
\definecolor{db}{RGB}{220, 230, 255}
\definecolor{lr}{RGB}{255, 240, 220} 
\definecolor{dr}{RGB}{255, 230, 200}
\definecolor{ld}{RGB}{240, 220, 255} 
\definecolor{dd}{RGB}{220, 200, 255}

\newcommand{\mhl}[2]{\sethlcolor{#1}\hl{#2}}

\title{\dataname: A Naturally Occurring Multimodal Dialogue Dataset for Team Reflection in Action During Clinical Operation}

\author{
 Naihao Deng$^{\text{\twemoji{peach}}}$ \quad
Kapotaksha Das$^{\text{\twemoji{lemon}}}$ \quad
Rada Mihalcea$^{\text{\twemoji{peach}}}$ \quad\\
{\bf Vitaliy Popov$^{\ast\text{\twemoji{peach}}}$ \quad}
{\bf Mohamed Abouelenien$^{\ast\text{\twemoji{lemon}}}$ \quad}\\
$^{\text{\twemoji{peach}}}$University of Michigan, Ann Arbor\quad
$^{\text{\twemoji{lemon}}}$University of Michigan, Dearborn\quad\\
{\tt \{dnaihao, zmohamed\}@umich.edu}
}

\begin{document}
\maketitle
\renewcommand{\thefootnote}{\fnsymbol{footnote}}
\footnotetext[1]{Both senior authors contributed equally to this work.}

\begin{abstract}
In clinical operations, teamwork can be the crucial factor that determines the final outcome.
Prior studies have shown that sufficient collaboration is the key factor that determines the outcome of an operation.
To understand how the team practices teamwork during the operation, we collected \dataname~from simulations of medical operations.
\dataname~includes the audio data and its transcriptions, the simulated physiology signals of the patient manikins, and how the team operates from two camera angles.
We annotate behavior codes following an existing framework to understand the teamwork process for \dataname.
We pinpoint three main characteristics of our dataset, including its label imbalances, rich and natural interactions, and multiple modalities, and conduct experiments to test existing LLMs' capabilities on handling data with these characteristics.
Experimental results show that \dataname~poses significant challenges to the existing models, inviting future effort on developing methods that can deal with real-world clinical data.
We open-source the codebase at \url{https://github.com/MichiganNLP/CliniDial}.\footnote[2]{Due to ethical considerations, the text data and video representations will be provided upon reasonable requests.}
\end{abstract}

\section{Introduction}

\begin{figure*}[t]
    \centering
    \includegraphics[width=\linewidth]{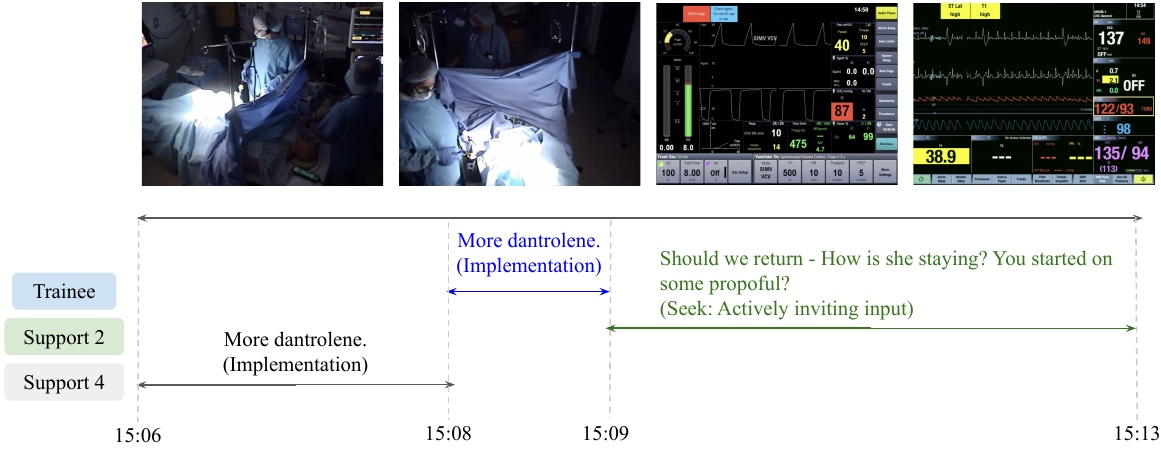}
    \caption{An example of the labeled dialogue in the simulated operation.
    Two cameras capture the scenes from two angles and two real-time monitoring systems provide the patient's physiological signals. 
    We only include the trainee and the two supports in this example, as they are the only three people speaking during this time frame.
    }
    \label{fig:dataset-example}
\end{figure*}

In clinical settings, teamwork is crucial for a successful operation, and effective team collaboration can improve the safety and well-being of the patients \cite{catchpole2008teamwork, weaver2010does, schmutz2019effective, rosen2018teamwork}. 
Failures in teamwork and communication among healthcare providers are a major contributing factor to the estimated 250,000 preventable deaths that occur in the U.S. each year \cite{rosen2018teamwork, makary2016medical}. 
Breakdowns in areas like leadership, situation awareness, decision-making and communication frequently underlie the many forms of preventable patient harm, including hospital infections, falls, diagnostic errors and surgical mistakes \cite{baker2005medical, herzberg2019association, keers2013causes}. 
There can be 58\% more deaths than expected due to insufficient collaboration \cite{knaus1986evaluation}.
Motivated by these statistics, in this paper we model the communication between team members as well as the data in the operation room to detect the effective steps and interactions needed for a successful procedure. 

To understand how teamwork unfolds in the operating room, we collected \dataname~from simulations of medical operations.
We collected the audio data, simulated physical signals from the patient manikins, as well as how the team operates from two camera angles.
We then annotated behavior codes based on a team reflection behavior framework \cite{schmutz2021reflection} to understand how the team members convey their objectives, strategies, and actions during the operation.
We provide initial analysis of our dataset, and lay out potential directions in using our dataset.
We hope researchers can leverage our dataset creatively, and propose methods to handle real-world clinical data.

In this paper, we pinpoint three main characteristics of \dataname, including its label imbalances, rich and natural interactions, and multiple modalities.
Corresponding to each feature, we design sets of experiments to investigate existing methods' ability to deal with such data, including the Large Language Models (LLMs) from GPT families and the open-source Llama families.
Experimental results show that \dataname~poses significant challenges to these methods.
In addition, we invite input from medical professionals to try to bridge the current NLP fields with the real-world applications they expect (\Cref{app-sec: medical-expectation}).

In summary, our contributions are two folds:
\begin{enumerate}[leftmargin=\parindent,align=left,labelwidth=\parindent,labelsep=0pt]
    \item We present \dataname, a naturally emerged multimodal dialogue dataset for team reflection during clinical operation.
    \item We evaluate our dataset against various existing methods with different setups and provide an analysis of their results.
    Our experimental results reveal that our dataset poses significant challenges to existing methods, urging methodology innovation in our NLP community.
\end{enumerate}


\section{How is \dataname~Different?}
\label{sec: difference}
Our real-world setting distinguishes \dataname~from existing datasets in various aspects.
First, there are significant \textbf{label imbalances} in the collected data.
Such label imbalances are less common in conventional NLP datasets where researchers have some levels of control over the data distribution by data filtering or downsampling.
However, since our dialogues occur naturally in the operation room, the interlocutors are not tasked to generate dialogues but rather to perform the clinical operation and take care of the ``patient'' as a team.
We do not pose any constraints on how the team communicate, and we observe that the amount of majority class labels significantly outmatches the minority class labels.
Second, there are \textbf{rich and natural interactions} between the team members.
Compared to the conventional dialogue benchmarks \cite{budzianowski-etal-2018-multiwoz} which typically contain 30 turns at most, the dialogue in our collected dataset contains 311 turns on average.
Third, there are \textbf{rich modalities} in the collected data. 
Compared to the conventional NLP datasets with text modality \cite{chen-etal-2021-dialogsum} or the conventional multimodal datasets which focus on vision and text modalities \cite{tapaswi2016movieqa, lei-etal-2018-tvqa, castro-etal-2022-wild}, the data we collect includes not only the dialogue, but also the corresponding audio, the operation views from two camera angles, and the physiological signals from the ``patient'' aligned for each timestamp.

\section{\dataname\ Dataset}
\label{sec: dataset-details}

\subsection{Data Descriptions}

\paragraph{Scenarios.} 
A team of board certified anesthesiologists  together with support staff is tasked with the intraoperative management of a 36-year-old female who is undergoing a minimally invasive surgery \footnote{The patient was diagnosed with acute cholangitis and is undergoing laparoscopic cholecystectomy}. 
This scenario takes place in a simulated operating room where we present a mannequin as the female patient and simulate her physiological signal changes from the backend.
Specifically, the patient develops malignant hyperthermia (MH; a rare complication of general anesthesia that could develop in any patient) as the simulated scenario progresses. Many healthcare providers lack sufficient clinical exposure to MH, potentially hindering their ability to recognize, treat, and manage these rare but severe cases effectively \cite{isaak2016review}.
We want to stress that this is not a real operation, and the intent is to train medical trainees in ``near-life'' surgical operations.

\paragraph{Roles.}
In the simulated operation, a confederate plays the role of the surgeon. 
The trainee who serves as the anesthesiologist is the main decision-maker \footnote{This is because malignant hyperthermia is a body's adverse reaction to an anesthetic.}. 
The support participants are also trainees who support an anesthesiologist.
\Cref{app-sec: scenario-details} provides additional details of the simulated operation and the roles of the team members.


\subsection{Labels}

\Cref{tab:label-examples} provides the definitions of each label and the corresponding examples.
Following \citet{schmutz2021reflection}, we include three labels of ``Seek'', ``Evaluate'' and ``Plan''.
As our data is sourced from clinical operations, we are interested in not only how the teams engage in reflection or diagnostic behaviors, but also how the team progresses from diagnostic actions to interventions or implementation actions.
Therefore, we assign an extra label ``Implement'' to such behaviors.
\Cref{app-sec:label-details} provides additional details for each label.
We describe the details of our annotation in \Cref{app-sec: annotation-details}.

\begin{figure}[t]
    \centering
    \begin{subfigure}[t]{0.9\linewidth}
        \centering
        \includegraphics[width=\linewidth]{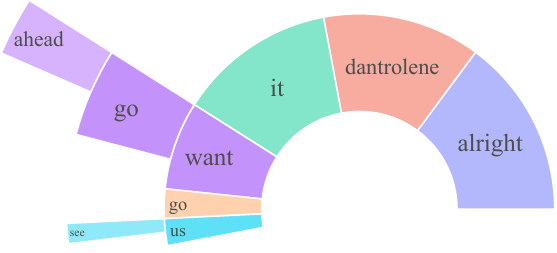}
        \caption{Distribution of words uttered.}
        \label{fig: support-sunburst}
    \end{subfigure}%
    \vspace{1em}
    \begin{subfigure}[t]{0.9\linewidth}
        \centering
        \includegraphics[width=\linewidth]{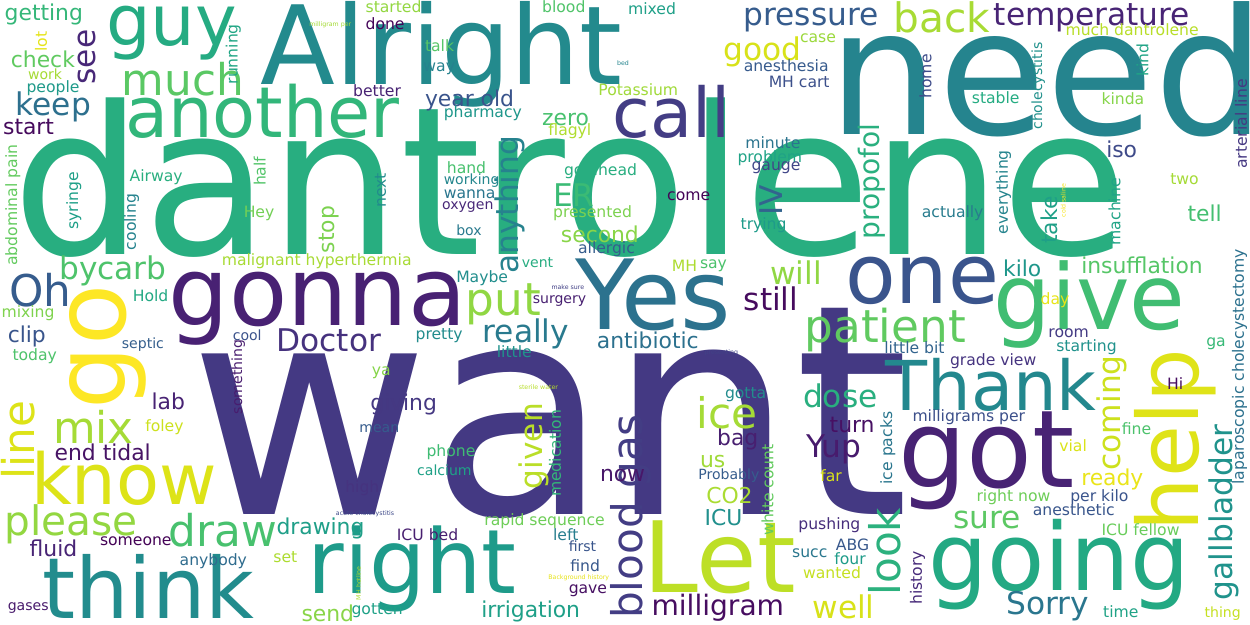}
        \caption{Word cloud for frequent words.}
        \label{fig:suppor-wordcloud}
    \end{subfigure}
    \caption{Distribution of words uttered (a) and word clouds for frequent words (b) by the support role.
    \Cref{fig:sunburst,fig:word-cloud} present the plots for all three roles.
    }
    \label{fig: support-plot}
\end{figure}

\begin{table}[t]
    \centering
    \small
    \begin{tabular}{crr}
    \toprule
    \multirow{2}{*}{General}& \# Sessions & 22 \\
    & \# Participants / Session & 6\\
    \midrule
    \multirow{4}{*}{Language} & \# Turns & 6.5k \\
    & \# Words & 49.9k \\
    & \# Turns / Session & 311 \\
    & \# Words / Session & 2.3k \\
    \midrule
    \multirow{3}{*}{Others} & Duration (min) / Session & 19 \\
    & \# Camera Angles & 2 \\
    & \# Physiological Signals & 9\\
    \bottomrule
    \end{tabular}
    \caption{Statistics of our collected dataset.}
    \label{tab:dataset-stats}
\end{table}

\subsection{Dataset Statistics and Analysis}
\label{subsec: dataset-analysis}
\Cref{tab:dataset-stats} provides the overall statistics of our collected dataset.
In total, there are 2,279 utterances in our dataset uttered by the support role, 1,808 by the surgeon role, and 2,576 by the trainee role.

\paragraph{Dataset example.}
\Cref{fig:dataset-example} provides an example of the annotated dialogue in the simulated operation.
As aforementioned, we have transcripts of different roles in the operation, together with camera views from two different angles, and the physiological signals of the patient mannequin.
We provide additional examples in \Cref{fig:more-examples} and dialogue snippets in  \Cref{tab:dialogue-examples} in \Cref{app-sec: dataset-analysis}.

\begin{figure}[t]
    \centering
    \begin{minipage}{0.48\textwidth}
        \centering
        \small
        \renewcommand{\arraystretch}{1.3}
        \begin{tabular}{crrrrr|r}
            \toprule
            Label & None & Seek & Eval & Impl & Plan  & All \\
            \midrule
            Overall & 3.7 & 1.3 & 0.8 & 0.6 & 0.3 & 6.9 \\
            Support & 1.0 & 0.4 & 0.2 & 0.3 & 0.1 & 1.9 \\
            Trainee & 1.2 & 0.4 & 0.4 & 0.2 & 0.1 & 2.2 \\
            Surgeon & 0.7 & 0.5 & 0.3 & 0.1 & 0.1 & 1.6 \\
            \bottomrule 
        \end{tabular}
        \captionof{table}{Total and role-specific label counts (in k).}
        \label{tab:label-dstr-num}
    \end{minipage}%
    \vspace{1em}
    \begin{minipage}{0.48\textwidth}
        \centering
        \begin{subfigure}[t]{0.8\linewidth}
            \centering
            \includegraphics[width=\linewidth]{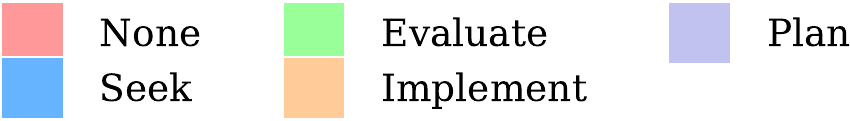}
        \end{subfigure}%

        \vspace*{1mm}
        \setcounter{subfigure}{0}
        \begin{subfigure}[t]{0.45\linewidth}
            \centering
            \includegraphics[width=\linewidth]{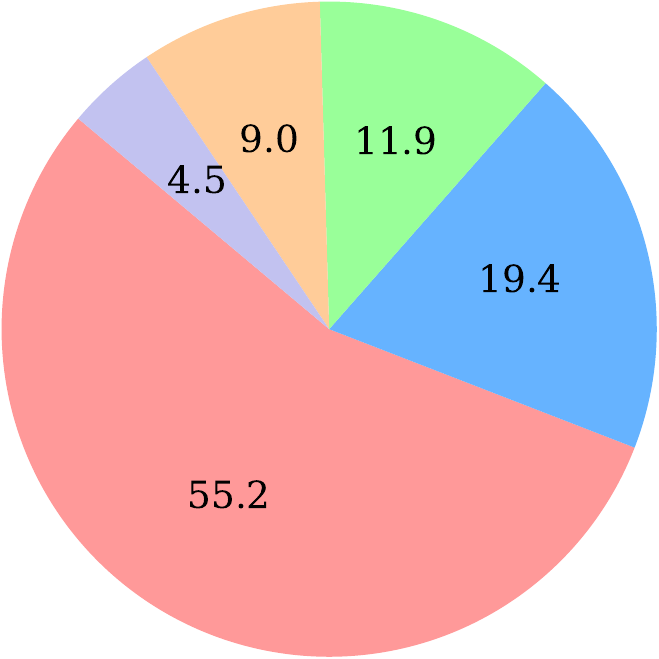}
            \caption{Overall.}
            \label{fig:overall}
        \end{subfigure}%
        ~
        \begin{subfigure}[t]{0.45\linewidth}
            \centering
            \includegraphics[width=\linewidth]{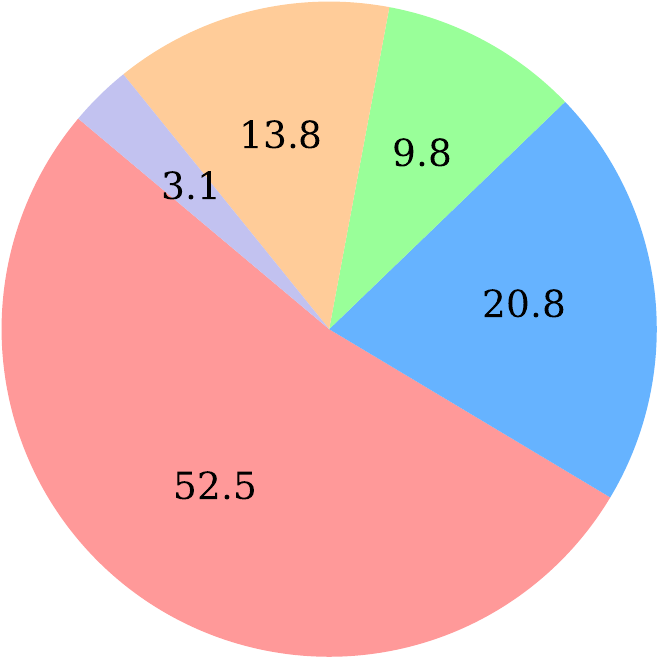}
            \caption{Support.}
            \label{fig:support}
        \end{subfigure}%

        \begin{subfigure}[t]{0.45\linewidth}
            \centering
            \includegraphics[width=\linewidth]{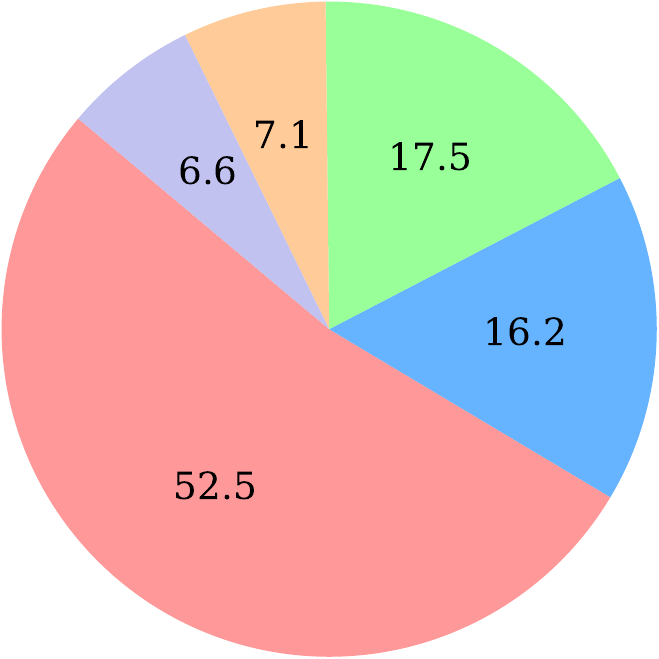}
            \caption{Trainee.}
            \label{fig:trainee}
        \end{subfigure}%
        ~
        \begin{subfigure}[t]{0.45\linewidth}
            \centering
            \includegraphics[width=\linewidth]{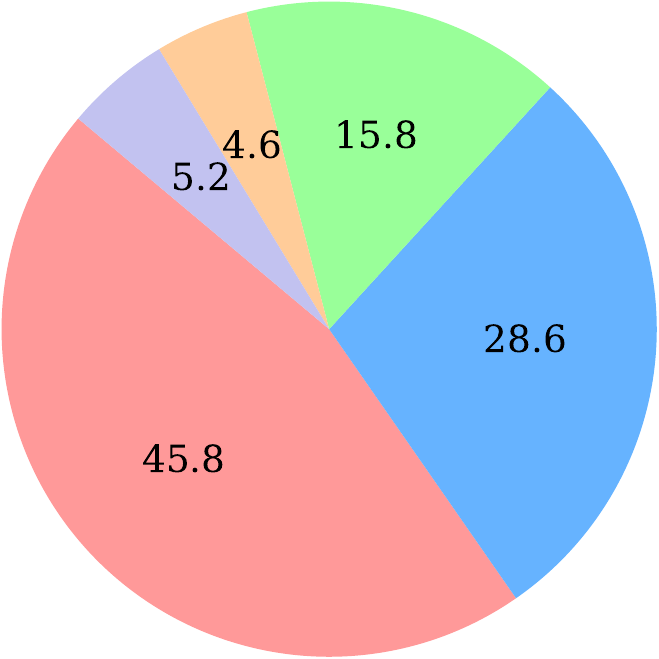}
            \caption{Surgeon.}
            \label{fig:surgeon}
        \end{subfigure}%
        \caption{Overall label distribution and role-specific label distributions.
        The numbers on the pie charts represent percentage values.}
        \label{fig:label-distr-pie}
    \end{minipage}
\end{figure}

\paragraph{Label distributions.}
\Cref{tab:label-dstr-num} provides the total and role-specific label counts.
\Cref{fig:label-distr-pie} provides the overall as well as role-specific label distributions.
Such role-specific label distributions reveal the internal collaboration and role-specific contributions during the operation.
We observe that, overall, the majority of labels are ``seek'' and ``evaluate,'' rather than ``implement'' or ``plan.''
This highlights the critical importance of communication and actively assessing the current situation during real-world operations.
Breaking it down by role, the surgeon role is most associated with the ``seek'' task, which accounts for 30.4\% of their labels. 
This indicates that, as the central figure in the operation, surgeons rely heavily on support and collaboration from other roles to fulfill their responsibilities effectively.
Additionally, the support role has the highest proportion of ``implement'' labels (13.7\%), which aligns with their primary function of providing assistance and executing essential procedures during the operation process.

\paragraph{Word analysis.}
\Cref{fig: support-plot} provides the token distributions and word cloud for the support role.
We provide these plots for all three roles in \Cref{fig:word-cloud,fig:sunburst} in \Cref{app-sec: dataset-analysis}.

In terms of the word distributions, we observe that trainees and surgeons often use ``thank you'', while supports often use ``alright'' (\Cref{fig:sunburst}).
This demonstrates the interactions happening during clinical operations, where there are such clues to acknowledge the actions conducted by others.
In addition, such phrase usages reflect the role difference.
Surgeons and trainees are the ones who need help from the support in the operation process, therefore they use ``thank you'' more often, while the support's primary job is to support others, therefore there is more of ``alright''.
Apart from these acknowledgment interactions, trainees often use the phrase ``CO2 up'' and surgeon often uses the phrase ``gallbladder out'', which relate to how they describe the situation or invite others' help during the operation process.
In terms of the frequent words, we observe many terminologies used by these roles.
For instance, both support and trainee roles frequently use the term ``dantrolene'' (\Cref{fig:word-cloud}), a medication primarily used to relax muscles, particularly in emergency settings \citep{krause2004dantrolene}.
Additionally, other medical terms appear frequently, such as ``abdomen'' and ``gallbladder'', which refer to anatomical structures; ``septic'', which relates to a patient's condition; and ``antibiotic'', which pertains to medication.

\paragraph{Camera view analysis.}
In addition, we provide a few qualitative observations based on videos captured from the two camera angles.
We observe the \textit{local focus on surgeon actions.}
Most operational actions occur within a localized region. For example, as seen in the screenshots in \Cref{fig:dataset-example}, the surgeon's body remains mostly stationary while manipulating tools during the operation.
In addition, we observe \textit{role-specific movement patterns.}
Supports and trainees tend to move around the room, creating distinct communication scenarios.
For instance, the support comes to the trainee to explain the background information in \Cref{subfig: information-seeking-example}.
In some scenes, the surgeon halts their operation to look around and communicate with colleagues. 
For instance, after talking to the trainee, the support moves to the doctor in the later scenes in \Cref{subfig: help-surgeon-example}. 
In other cases, the surgeon continues working while colleagues stop to seek information from them.
These contrasting behaviors can be interesting clues to understanding role-specific dynamics, and we plan to include additional examples of such interactions in the final version.
Our dataset poses \textit{unique visual challenges in operation settings.}
The people in the videos are different from those typically seen in everyday video datasets. 
As shown in \Cref{fig:dataset-example}, participants wear uniforms, hats, and facial masks, which obscure facial expressions. 
For example, in \Cref{fig:dataset-example}, a human observer might easily infer that the surgeon is frowning while staring at a monitor, even though most facial features are obscured. However, such subtleties can pose challenges to the VLMs. We would be happy to learn if you have suggestions for methods from the vision community that could help us further analyze the visual information in our dataset.

\Cref{app-sec: dataset-info} provides additional information for the dataset as well as the physiological signals included.
We apply ten-fold cross-validation on our dataset and report the average macro and micro F1 scores in the following setups.
For each fold, we use 17, 2, and 3 sessions for training, validation, and testing, respectively.


\subsection{Potential Usage of \dataname}

In this paper, we present three case studies scrutinizing existing LLMs' capabilities on handling domain-specific data with specific characteristics in \Cref{sec: class-imblanace,sec: conversation-nature,sec: multimodality}.
We include two potential usages of \dataname\ and encourage future research in using our dataset in creative ways.

\paragraph{Testing the Effectiveness of Existing Methods.}
We present 6.9k examples of annotated examples in \dataname.
In this paper, we take the first step to test various methods including LLMs' capabilities in handling clinical data, especially on data with imbalanced class distribution, conversational nature, and multiple modalities.
We highlight that our annotated dataset can be a valuable source to investigate the innate ability of LLMs in handling real-world clinical data.

\paragraph{Understanding the Interaction Mechanisms in Clinical Setups.}
As demonstrated in \Cref{subsec: dataset-analysis}, our dataset presents a valuable source of interactions across different roles that happen in clinical operations.
There is rich domain jargon involved such as ``gallbladder'', ``dantrolene'', and phrases that are specific to the clinical operation setup, such as ``CO2 up'', etc.
Moreover, there are interesting phrase usage patterns during such interaction processes.
For instance, surgeons and trainees use ``thank you'' often, while supports use ``alright'' often.
Such language use patterns reflect what tasks each role carries out during the clinical operations, and how one role reacts to the requests or actions of the others.
Therefore, the dialogue interaction in \dataname\ can facilitate future research on understanding the interaction mechanisms in the clinical operation.

\paragraph{Incorporated in LLM's Training Loop.}


When we collect the dataset, we have included the timestamps for different modalities as shown in \Cref{fig:dataset-example}. 
Such timestamps can map information across different modalities. 
For instance, given certain frames from the videos, we can pinpoint the sentence uttered by the surgeon and the supports correspondingly. 
This mapping can enable LLM training objectives such as masking information in one modality and then asking LLMs to predict the missing signals based on the information from all the remaining modalities. 
We highlight that despite the difficulty of the data collection process, \dataname\ includes 6.5k turns and 49.9k words in total.
Though such amount of data may not be sufficient for pre-training from the scratch, researchers may adopt our dataset for continual pre-training to facilitate models in clinical domains.
In addition, we present 6.9k examples annotated with labels, which can serve as a valuable source in models' supervised fine-tuning stage.

In the following sections, we present three case studies revealing the challenges for existing methods including LLMs in dealing with our dataset.


\section{Charactersitic I: Imbalanced Class Distribution}
\label{sec: class-imblanace}

\begin{figure}[t]
    \centering
    \includegraphics[width=0.9\linewidth]{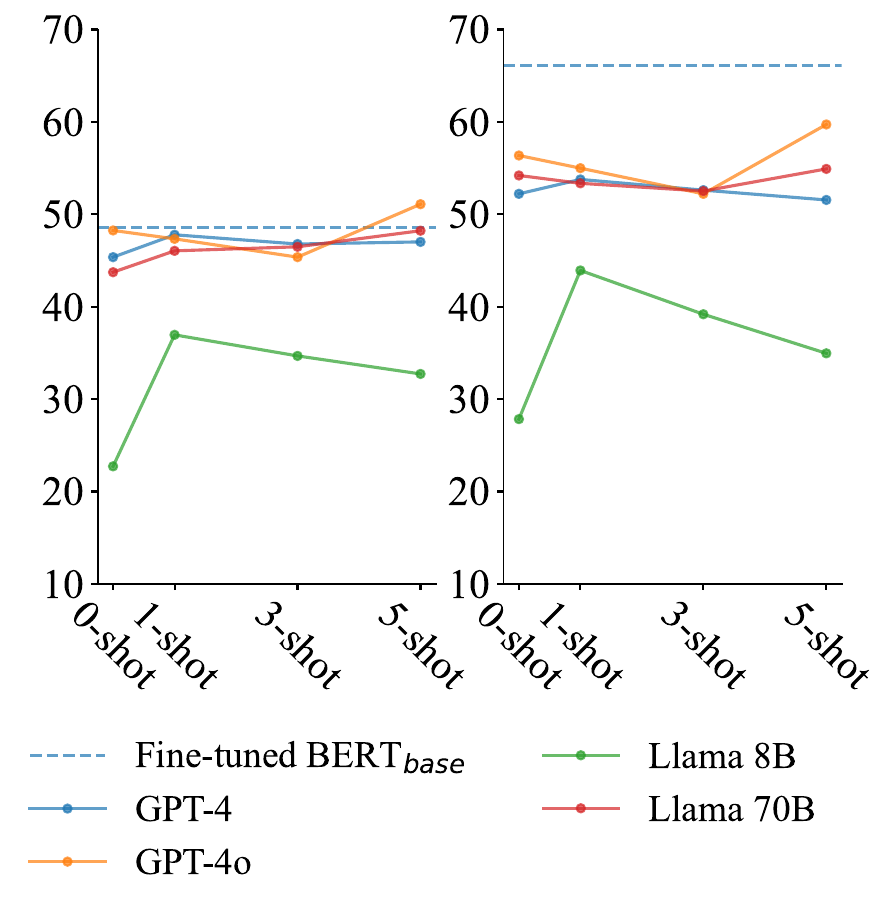}
    \caption{Comparison of macro F1 scores (F1 scores averaged by class, on the left) and micro F1 scores (F1 scores averaged by instances, on the right) versus number of demonstrations (number of shots).
    We compare both scores for the fine-tuned BERT$_{\text{base}}$ model, 0-shot and few-shot prompting for LLMs.}
    \label{fig:shot-all}
\end{figure}


Here we constrain our study within the text domain for handling label imbalance.

\subsection{Evaluation Setups}

We directly fine-tune a BERT base \cite{devlin-etal-2019-bert} model to learn directly from the skewed data.
In addition, we prompt Llama 3 8B and 70B models (abbreviated as Llama in figures) and GPT-4 and GPT-4o with and without demonstrations.
Specifically, we use the following prompt to provide the list of all possible labels:
\begin{mybox}{Prompt}
In the classification task, there are 5 labels: [Seek, Evaluate, Plan, Implement, None].

Here are the details for each label:

[Description of each label]

Fill in the blanks:

Output in the format of \{

\quad``Sentence'': <SENTENCE>,

\quad``Label'': ,

\}
\end{mybox}
In the few-shot settings, we provide corresponding examples along with the label definitions to the models.
\Cref{tab:label-examples} in \Cref{app-sec:label-details} provides labels, their corresponding definitions, and examples.
We adapt the definition from the corresponding information provided in the annotator guidelines.
``Examples'' provides the demonstrations of examples sourced from the dataset.
We use this prompt for the prompting experiments in \Cref{sec: class-imblanace,sec: conversation-nature,sec: multimodality}.
\Cref{app-sec: more-model-details} provides additional baseline models and their results.
We extract the label from the JSON output of the model to calculate the macro F1 scores (F1 scores averaged by class) and the micro F1 scores (F1 scores averaged by instances).


\subsection{Discussions}

\paragraph{Tuning-based method.}
\Cref{fig:shot-all} compares the F1 scores averaged by class (macro F1 scores) and F1 scores averaged by instances (micro F1 scores).
Though the fine-tuned BERT$_\text{base}$ model can achieve the highest micro F1 score of 66.6\%, it yields the macro F1 score of 48.6\%, which is much lower compared to its micro F1 score, and is comparable to GPT-4o's macro F1 score at 0-shot (48.3\%) or 5-shot (51.1\%).
This suggests that tuning-based methods bias the model to better learn the majority class, while the LLMs with a few demonstrations from each class do not suffer from the performance disparity between the macro and micro F1 scores. 

\paragraph{Prompting-based method.}
There is a significant performance boost for Llama 8B from 0-shot, achieving a macro F1 score of 22.7\% to 1-shot, achieving a macro F1 score of 37.0\%, suggesting even a single example can guide smaller LLMs to better reason.
However, when we increase the number of demonstrations, the Llama 8B model experiences a performance decline, from a macro F1 score of 37.0\% at 1-shot to 34.7\% at 3-shot and 32.7\% at 5-shot.
In contrast, there is slight performance improvement for the Llama 70B model when we increase the number of demonstrations, from a macro F1 score of 43.7\% at 0-shot, to 46.0\% at 1-shot, 46.5\% at 3-shot and 48.2\% at 5-shot.
We attribute such a phenomenon to the limited innate capabilities of Llama 8B model, as the smaller scale model may not capture the underlying knowledge from a few demonstrations, instead it may be distracted by the longer input when we increase the number of demonstrations.
Moreover, there is only slight performance improvement for Llama 70B and GPT models when we increase the number of demonstrations.
For the Llama 70B model, its macro F1 score improves from 46.0\% at 1-shot to 48.2\% at 5-shot.
For the GPT-4 model, its macro F1 score remains around 47\% when we increase the shot number from 1 to 5, while for GPT-4o model, its macro F1 score improves from 47.3\% at 1-shot to 51.1\% at 5-shot.
We hypothesize that the real-world nature of our dataset leads to diverse dialogue patterns, making a few demonstrations insufficient for the model to cover all scenarios.

\section{Charactersitic II: Conversational Nature}
\label{sec: conversation-nature}
As discussed in \Cref{subsec: dataset-analysis}, \dataname\ involves rich interactions among people where they actively communicate information in the operation process.
Hence, an ideal model would leverage the context information of the interaction.

\subsection{Evaluation Setups}
\begin{figure}[t]
    \centering
    \includegraphics[width=0.9\linewidth]{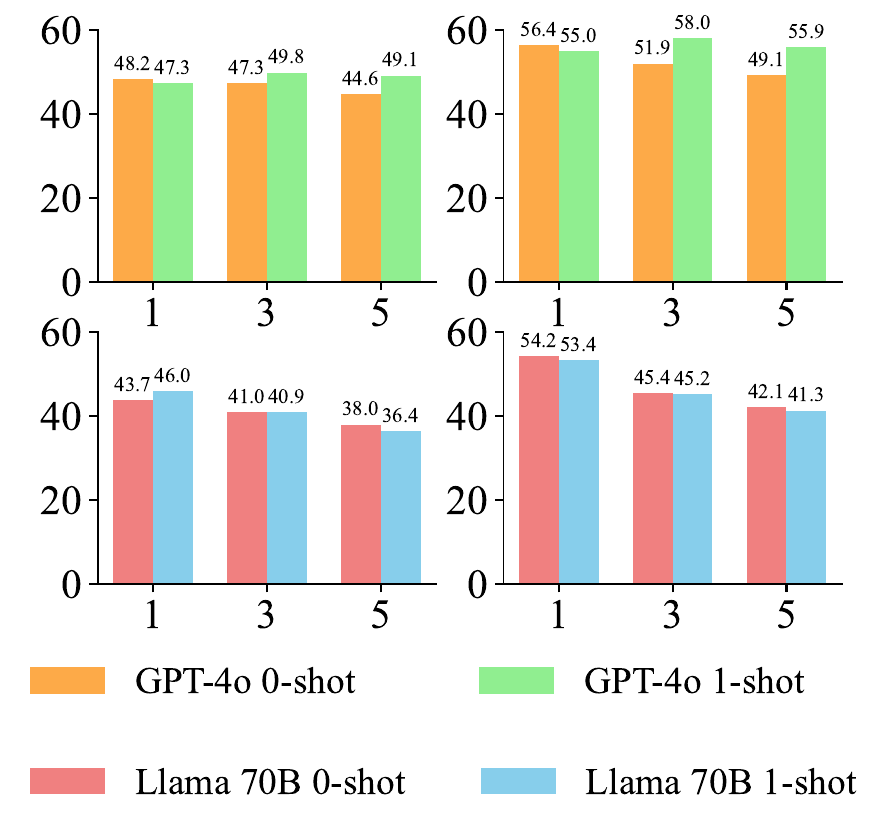}
    \caption{Comparison of macro F1 scores (F1 scores averaged by class, on the left) and micro F1 scores (F1 scores averaged by instances, on the right) versus the context size (x-axis).
    For instance, ``3'' on x-axis represents a context of size ``3'', where we include one turn both before and after the current turn in our prompt to the LLM.}
    \label{fig:context-all}
\end{figure}

We take the best performed closed-source LLM, GPT-4o, and the best performed open-source LLM, Llama 70B from \Cref{fig:shot-all}.
We then prompt them with one turn both before and after the current round (context size of 3 in \Cref{fig:context-all}) or two turns before and after the current turn (context size of 5 in \Cref{fig:context-all}).
Specifically, we insert the following additional prompt into the prompt we use in \Cref{sec: class-imblanace} after the label descriptions:
\begin{mybox}{Additional Prompt}
For the dialogue:

<CONTEXT BEFORE>

<ROLE>: <SENTENCE>

<CONTEXT AFTER>
\end{mybox}
In the prompt, <CONTEXT BEFORE> and <CONTEXT AFTER> correspond to the turns before the current utterance and the turns after, and the model needs to assign a label for <SENTENCE>.
In both situations, we report the performance by providing no demonstration (0-shot) or a single demonstration (1-shot).


\subsection{Discussions}
\Cref{fig:context-all} reports the performance comparison across different settings.
For GPT-4o, we observe a performance boost when we include the interactions.
For instance, under the 1-shot setup, the GPT-4o model's macro F1 score improves from 47.3\% to 49.8\% and micro F1 scores improve from 55.0\% to 58.0\% when we increase the context size from 1 to 3.
However, when we further increase the context size to 5, it suffers a performance decline compared to the context size of 3, but still outperforms the case when the context size is 1.
This indicates that context can help models better reason the target sentence, but when too much context is provided, the information may be diluted and is less helpful.
In contrast, providing demonstrations and increasing context size negatively impact Llama 3's performance.
Under the 1-shot setup, Llama 3's macro F1 score drops significantly, from 46.0\% to 40.9\% when the context size increases from 1 to 3, and further to 36.4 when the context size increases from 3 to 5. 
We attribute this performance decline to the increased input length. 
On average, including context information and one demonstration results in an input length of approximately 1,000 tokens per example, utilizing one-eighth of Llama 3's 8k context window.
We hypothesize that Llama 3, with its smaller context window, struggles to process such long inputs effectively, consistent with findings by \citet{he2024can}. 
In contrast, GPT-4o, equipped with a much larger context window of 128K, is better suited to handle input lengths of this magnitude.

\section{Charactersitic III: Multimodality Beyond Text and Vision}
\label{sec: multimodality}

\begin{figure}[t]
    \centering
    \includegraphics[width=\linewidth]{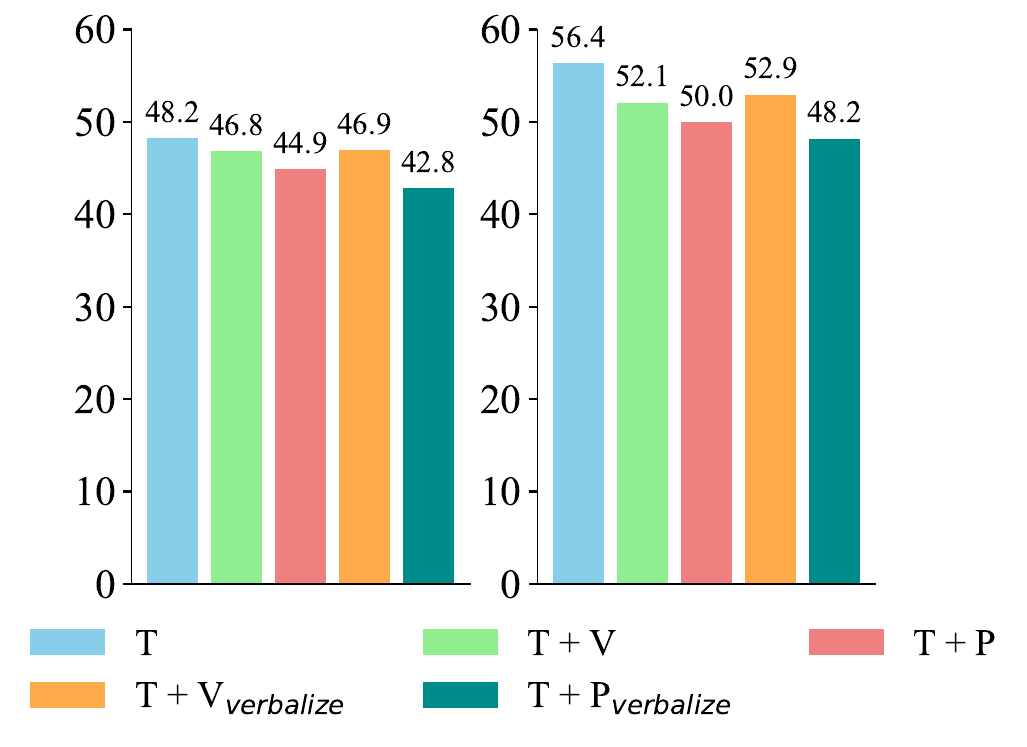}
    \caption{Comparison of macro F1 scores (F1 scores averaged by class, on the left) and micro F1 scores (F1 scores averaged by instances, on the right) when we pass in different modalities.
    ``T'' standars for text-only, ``V'', ``P'' standard for visual signals and physiology signals, respectively.
    ``T + V$_\text{verbalize}$'' and ``T + P$_\text{verbalize}$'' stand for verbalizing the content by GPT-4o first, and then pass the text description with the other instructions to the GPT-4o model.}
    \label{fig:multimodal-results}
\end{figure}


\begin{table*}[ht]
    \centering
    \small
    \begin{tabular}{cp{13cm}}
        \toprule
       \multirow{6}{*}{V\textsubscript{verbalized}}  &  The video frames show a surgical operation taking place in an operating room. In both frames, the surgical team, including at least two members dressed in blue sterile gowns and gloves, is actively engaged in a procedure. The patient is lying on the operating table, partially covered by surgical drapes. A bright surgical light illuminates the operative field, and an anesthesia machine and monitoring equipment are visible nearby. The scene appears to be well-organized, with a focus on maintaining a sterile environment.\\
         \midrule
       \multirow{3}{*}{P\textsubscript{verbalized}}  & The patient is on mechanical ventilation (SIMV VCV mode) with normal ventilator settings. However, there is a warning for high EtCO$_2$ (87 mmHg), indicating hypoventilation or CO$_2$ retention. Oxygen saturation is 99\%, and heart rate is \mhl{ld}{64 bpm}—stable but concerning for rising CO$_2$ levels. \\
       \bottomrule
    \end{tabular}
    \caption{An example of V\textsubscript{verbalized} and P\textsubscript{verbalized} corresponding to the scenes in \Cref{fig:dataset-example}.
    The model incorrectly assigns the value of \mhl{ld}{64} to heart rate, which is the Et value instead.
    }
    \label{tab:example}
\end{table*}

\subsection{Evaluation Setups}
We evaluate the GPT-4o model, a multimodal end-to-end LLM with different modalities as the input, including feeding pure text (T), text and the operation video from two angles (+V), text and the physiology signals (+P).
In addition, we try to let GPT-4o first verbalize what happens in the camera views or the physiological signals by prompting:
\begin{mybox}{Prompt for Verbalization}
The video frames show the [operation scene for/ physiological signal changes from] this patient. Please describe what happens in a few sentences.
\end{mybox}
After we acquire the verbalized descriptions, we feed the following description after the label definitions we use in the prompt in \Cref{sec: class-imblanace}.
\begin{mybox}{Description}
Here is the description for what happens on the scene: <DESCRIPTION>
\end{mybox}

\subsection{Discussion}
From \Cref{fig:multimodal-results}, we can see that GPT-4o fails to leverage the visual or the physiological signals effectively.
When we add the visual scene input directly, the model's macro F1 score decreases from 48.2\% (text-only) to 46.8\%, and when we add the screenshot of the physiological signals directly, the model's macro F1 score decreases to 44.9\%.
This demonstrates that domain-specific data from modalities other than text pose significant challenge to advanced LLMs like GPT-4o.
Specifically, for the operation scenes and physiological signals in \Cref{fig:dataset-example}, while a trained medical professional may understand the situation and read the monitor correctly, we hypothesize that the dark scene of the camera and the strong light focusing on the patient's abdomen area may pose challenges to LLMs in their reasoning process.
Moreover, since there would not be too many scenes online corresponding to physiological signals, GPT-4o may have not encountered such data in its pre-training process, leading to its limited capabilities to process it. 
When the physiological signals are verbalized instead of presented as images, GPT-4o experiences an additional 2\% performance drop.
This highlights the challenges of incorporating physiological data into model's reasoning process, as the errors accumulate during the two-step process of first verbalizing the data and then reasoning over the verbalized context.
In contrast, when we verbalize the visual scenes, GPT-4o performs comparably (46.9\% versus 46.8\% for macro F1 score) to when we pass the visual scenes directly.
This indicates that GPT-4o can better handle text-based representations of visual information than it can with text-based representations of physiological signals.
We attribute this to GPT-4o's lack of domain-specific knowledge, unlike visual scenes, interpreting physiological signals requires more specialized expertise.

\paragraph{Case Study.}
We present an example of V\textsubscript{verbalized} and P\textsubscript{verbalized} corresponding to the scenes in \Cref{fig:dataset-example} in \Cref{tab:example}.
For V\textsubscript{verbalized}, the model describes the scenes accurately.
However, the model hallucinates in the P\textsubscript{verbalized}. 
In \Cref{tab:example}, 64 is not the heart rate, instead, it is the Et value for the gases. 
We believe that this is an important direction for future research to make models correctly reason the situations, especially in a high-stakes domain such as clinical operations.


\section{Related Work}

\paragraph{Multimodal Datasets.}
Recent years have witnessed significant advancement of multimodal large language models (MLLMs) \citep{achiam2023gpt, liu2024visual} that typically involves vision and text capabilities.
These MLLMs have demonstrated impressive performance on various visual benchmarks such as visual recognition \citep{zhang2024vision}, video understanding \citep{xu-etal-2021-vlm}, 3D understanding \citep{hong20233d} and beyond.
Researchers have proposed various vision and text benchmarks to investigate the capabilities of these MLLMs, including captioning tasks such as MS-COCO \citep{lin2014microsoft} and Flickr30K \citep{plummer2015flickr30k} for image captioning, and MSVD \citep{chen2011collecting} for video captioning, question answering tasks such as VQA \citep{antol2015vqa} for image question answering and MSVD-QA \citep{xu2017video} for video question answering.
Recently, there is a shift of interest in proposing more nuanced and culturally diverse benchmarks.
For instance, WildQA proposes video QA dataset on scenes in the wild \citep{castro-etal-2022-wild}, Ego4D proposes various visual tasks from the egocentric viewpoint \citep{grauman2022ego4d}, CVQA investigates into the culturally diverse multilingual visual question answering \citep{romero2024cvqa}.
In addition, researchers have proposed datasets involving other modalities, such the Touch and Go dataset on tactile \citep{yang2022touch}, MMAU on audio understanding \citep{sakshi2024mmau}.
To the best of our knowledge, we are the first to propose a dataset that includes the physiological signals.
Moreover, we provide timestamps for the utterances, which allows researchers to align the text data with video frames and the physiological signals.

\paragraph{Datasets in Clinical Domains.}
There has been interdisciplinary research between NLP and clinical or medical domains \citep{spasic2020clinical}.
For instance, researchers have leveraged natural language generation methods to generate medical reports or summaries \citep{song-etal-2020-summarizing, papadopoulos-korfiatis-etal-2022-primock57, ben-abacha-etal-2023-empirical}, understanding the medical consultant process \citep{chen2023benchmark}.
However, most of these existing datasets focus on the consultant process in the clinical setup.
We highlight that \dataname\ focuses on the conversation during clinical operations, which possess significant domain-specific features as discussed in \Cref{subsec: dataset-analysis}.
We hope \dataname\ can facilitate future NLP research into understanding the complex clinical operation scenarios.

\section{Conclusion}

In this paper, we introduced \dataname, a naturally emerged multimodal dialogue dataset collected from clinical operations. 
Unlike existing benchmarks, \dataname\ addresses real-world complexities such as imbalanced label distributions, rich team interactions, and multiple data modalities.
Through studies on three key characteristics of our dataset, we found that the best-performing model achieves a macro F1 score of only 51.09, indicating significant room for improvement.
This performance suggests that existing methods struggle on CliniDial, particularly in handling imbalanced class distributions, leveraging conversational context, and integrating domain-specific multimodal signals.
These limitations highlight the gap between existing NLP methods and the demands of real-world clinical applications.
We hope \dataname\ can bridge the gap between advancements in our community and real-world clinical applications. 
We encourage future research efforts to develop domain-adaptive NLP techniques, improve multimodal fusion strategies, and eventually address the challenges of real-world applications.

\section*{Limitations and Future Directions}

\paragraph{Simulation Setup.}
The clinical operation described in the study is simulated, so it is likely that the dialogue between the anesthesiologists and support staff lacks the sense of urgency present in a real medical setting. 
In real-life clinical environments, time pressure, high-stress situations, and the need for quick decision-making usually shape the communication dynamics. 
This distinction is important to consider when analyzing the dialogue, as the lack of urgency might influence both the content and the tone of communication.

However, to the best of our knowledge, we are the first to study such a medical operation scenario, even in a simulated operation process. 
In fact, it would be nearly impossible to collect the real emergency operation recordings due to ethical and legal considerations.
These simulations are the typical training that medical professionals rely on, and to the best of our knowledge, the best possible way to collect such data. 

\paragraph{Scope of the Data.}
We want to emphasize the difficulty of setting up the real-world clinical operation environment, recruiting people to participate, collecting the data.
Although the dataset is collected mainly on 22 clinical operation sessions, we note that there are 6.5k turns and 49.9k words in total in \dataname.

\paragraph{Scope of the Analysis.}
We provide various analyses on our dataset in \Cref{sec: dataset-details} and highlight how our dataset is different from the existing benchmarks in \Cref{sec: difference}.
In addition, we discuss the potential future directions that researchers may explore in \Cref{sec: dataset-details}.
Furthermore, we study the three characteristics of our dataset and provide the performance of popular NLP methods with respect to each of them.
However, due to the scope of this study, we cannot evaluate every possible method and would like to invite future efforts on a comprehensive evaluation of NLP methods on clinical data.
For instance, our dataset can be leveraged to answer questions such as ``are there sequences of labels that occur frequently in the corpus?''
We encourage future efforts on a more in-depth exploration that might reveal underlying structures or recurring communication patterns in the dialogues between anesthesiologists and support staff, and provide a richer understanding of the linguistic dynamics.

\paragraph{Scope of the Experiments.}
We encourage future efforts to investigate the low F1 scores for the existing LLMs. 
For instance, prompting methods such as chain of thought (CoT) prompting could be tested to check whether they could enhance the LLM's performance and lead to a higher F1 score, which can lead to a more reliable approach for analyzing clinical dialogues.
In this paper, we did not include analyses of the audio setting.  
Audio characteristics can provide additional insights into the emotional state, stress, urgency, or intent behind the spoken words, offering a better understanding of what's really going on.
We leave the exploration of models fine-tuned with medical expertise to future study.
To the best of our knowledge, there are no specific LLMs targeting clinical operation setup.

\section*{Ethics Statement}
We note that the study was approved by the Institutional Review Board.
Since the data from the two cameras may reveal the identity of the team, we may not release the camera data.
We are considering to release an anonymized version of the dialogue transcription to facilitate future research on clinical NLP.
We expect researchers to continue building new algorithms and methods on top of this clinical dataset.

\section*{Acknowledgments}
This study is based upon work partially supported by the National Science Foundation under Grant IIS-2202451 and grant IIS-2306372. Any opinions, findings, and conclusions or recommendations expressed in this material are those of the authors and do not necessarily reflect the views of the National Science Foundation.


\bibliography{anthology,custom}

\appendix

\newpage
\section{Label Details}
\label{app-sec:label-details}

\begin{table}[t]
    \centering
    \small
    \begin{tabular}{cl}
        \toprule
        Labels &  Behavior Subcodes\\
        \midrule
        \multirow{2}{*}{Seek} & Actively inviting input\\
        & Expressing uncertainty \\
        \cmidrule{1-2}
        \multirow{4}{*}{Evaluate} & Stating a working hypothesis \\
        & Recapping \\
        & Explicitly assessing the situation \\
        & Reasoning \\
        \cmidrule{1-2}
        Plan & Stating plans and priorities \\
        \midrule
        \multirow{2}{*}{Implement} & Stating one's ongoing actions\\
        & Designating tasks \\
        \bottomrule
    \end{tabular}
    \caption{Behavior subcodes corresponding to each of our labels.
    We follow the definition from \citet{schmutz2021reflection} to determine the subcodes for ``Seek'', ``Evaluate'', and ``Plan''.
    We add another label of ``Implement'' given the characteristics of our data source.}
    \label{tab:label-brief-intro}
\end{table}

\begin{table*}[t]
    \small
    \centering
    \begin{tabular}{cp{42em}}
        \toprule
        Label & Seek \\
        \grayc Definition & \grayc All statements that request information from the team about the current event and invite team members to provide information and share ideas; Inquiring for further information. 
        Or expressions of uncertainty with an implicit invitation to share information. Tone of voice; Content of what is being said (questioning the information; Unsure). 
        In response to clarify something. \\
        Examples & Is there anything we are missing? Is there anything else we should be doing? \\
        & What is the plan afterwards? \\
        \midrule
        Label & Plan \\
        \grayc Definition & \grayc Laying out the course of action for the next few minutes.  Needs to contain at least 2 actions to show a sequence of actions. \\
        Examples & Once MH is recognized: Going to stop the agent and go up on flows. \\
        \midrule 
        Label & Evaluate \\
        \grayc Definition & \grayc Clear formulation of a working hypothesis or diagnosis about the current situation, or various pieces of information are brought together and a summary is provided; Recapping lab results (does not have to be new information); Providing an explicit judgment for something, give value to a certain process, information or strategy ...\\
        Examples & That's a nasty gallbladder. \\
        \midrule
        Label & Implement \\
        \grayc Definition & \grayc Stating one's ongoing actions or designating tasks. \\
        Example & Yes. Yes. Pushing. Pushing. Pushed. \\
        \midrule
        Label & None \\
        \grayc Definition & \grayc None of the other labels apply here. \\
        Examples & Okay. \\
        \bottomrule
    \end{tabular}
    \caption{Examples of the labels, their definitions, and corresponding utterances in our dataset.
    We omit part of the definitions for the label ``Evaluate''. 
    \Cref{app-sec:label-details} provides additional details of each label.
}
    \label{tab:label-examples}
\end{table*}

\Cref{tab:label-brief-intro} provides an overview of the behavior subcodes for each label. 
\Cref{tab:label-examples} provides the comprehensive definition and examples corresponding to each label.

\paragraph{Seek} includes:
\begin{itemize}
    \item the action of actively inviting the team members to provide information and share ideas about the current event.
    \item expressing uncertainty with an implicit invitation to share information.
\end{itemize}

\paragraph{Evaluate} includes:
\begin{itemize}
    \item a clear formulation of a working hypothesis or diagnosis about the current situation.
    \item bringing together various pieces of information and providing a summary.
    \item providing an explicit judgment, giving value to a certain process, information, or strategy. 
    This can be the process of evaluating information that has been gained through seeking information. 
    \item explaining why certain things are more important, or why a specific behavior needs to be done.
\end{itemize}

\paragraph{Plan} refers to laying out the course of action for the next few minutes that needs to contain at least two actions.

\paragraph{Implementation} refers to stating the member is conducting the task or delegates a task to another team member.

\section{Scenario Details}
\label{app-sec: scenario-details}

The role of primary anesthesiologist was played by one of the course participants. The surgeon and secondary anesthesiologist (assistant) were
played by other course participants. The role of surgeon served as a confederate along
with the course instructors. The scenario begins with the primary anesthesiologist
taking over the case from one of the course instructors. The patient is receiving general
anesthesia and the procedure has already begun. The procedure is complicated by
surgical difficulties resulting in the surgeon requesting additional muscle relaxants and
increased insufflation pressures. There is also concern that the patient is developing
sepsis given the significant gallbladder infection.
The patient develops malignant hyperthermia (MH) as the simulated scenario progresses. 
The primary anesthesiologist must recognize this and begin appropriate treatment. 
Treatment algorithms for MH are well-known and broadly available
(Hopkins et al., 2020; Rosenberg et al., 2020). Definitive treatment includes stopping
the triggering agents, administering dantrolene, and supportive care.

\section{Dataset Information}
\label{app-sec: dataset-info}

The total number of anesthesiologists studied was 22; 15(68\%) males and 7(32\%)
females. As part of the Maintenance of Certification in Anesthesiology (MOCA©),
anesthesiologists who were board certified after 2000 were required to participate in a
simulation course at a simulation center.
The participants were board certified anesthesiologists who
attended a simulation course at a midwestern academic medical center over a 5 year
period. Date of initial certification was obtained from the American Board of
Anesthesiologists (ABA) Physician Directory. The study was approved by the
Institutional Review Board.

\subsection{Physiological Signals}

The physiological signals in our dataset include:

\paragraph{SpO2} refers to Peripheral Oxygen Saturation which measures the oxygen saturation level in the blood.
Such signal is typically measured through a pulse oximeter.

\paragraph{ECG II} refers to Electrocardiogram Lead II which represents the electrical activity of the heart as measured by electrodes placed on the body.

\paragraph{APB} refers to Arterial Blood Pressure which represents the pressure exerted by blood on the walls of the arteries during the cardiac cycle.

\paragraph{HR} refers to Heart Rate which indicates the number of heartbeats per minute.

\paragraph{NIBP} refers to Non-Invasive Blood Pressure which measures blood pressure without the need to insert instruments into the body.

\paragraph{Temperature} represents the body's temperature and is often measured using a thermometer.

\paragraph{Respiratory Waveform} represents the pattern of inhalation and exhalation.

\paragraph{CO$_{2}$} means Carbon Dioxide which typically refers to end-tidal CO$_{2}$, which represents the concentration of carbon dioxide at the end of an exhaled breath.

\paragraph{IBP} refers to Invasive Blood Pressure which measures blood pressure using invasive techniques, typically involving a catheter inserted into an artery or vein.

\subsection{Annotation Details}
\label{app-sec: annotation-details}

Two researchers coded six out of 22 randomly selected data files. 
The researchers discussed findings and resolved discrepancies through the process of social
moderation. 
They achieved a Cohen’s kappa score of 0.73 on the six files.
The two researchers then independently annotated the remaining dataset.

\section{Dataset Analysis}
\label{app-sec: dataset-analysis}

\begin{figure*}[th]
    \centering
     \begin{subfigure}[t]{0.9\linewidth}
        \centering
        \includegraphics[width=\linewidth]{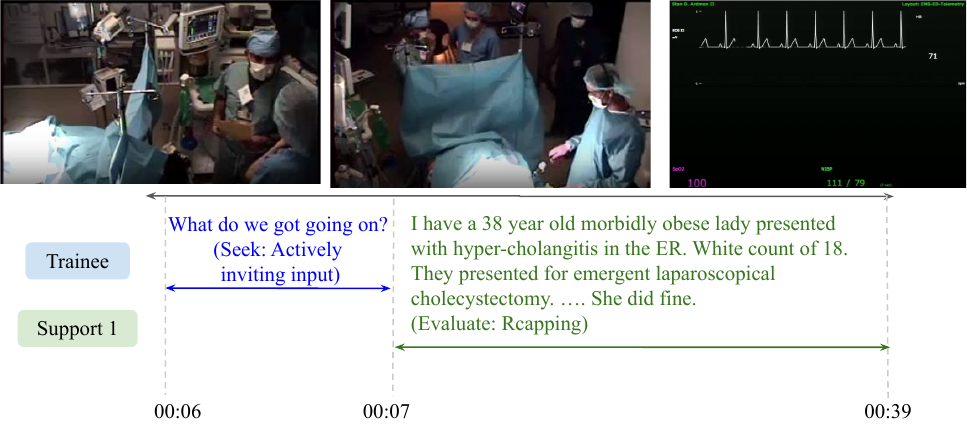}
        \caption{An example happens at the beginning of the operation, when the trainee seeks background information and the support provides such information to the trainee.}
        \label{subfig: information-seeking-example}
    \end{subfigure}%
    \vspace{2em}
    \begin{subfigure}[t]{0.9\linewidth}
        \centering
        \includegraphics[width=\linewidth]{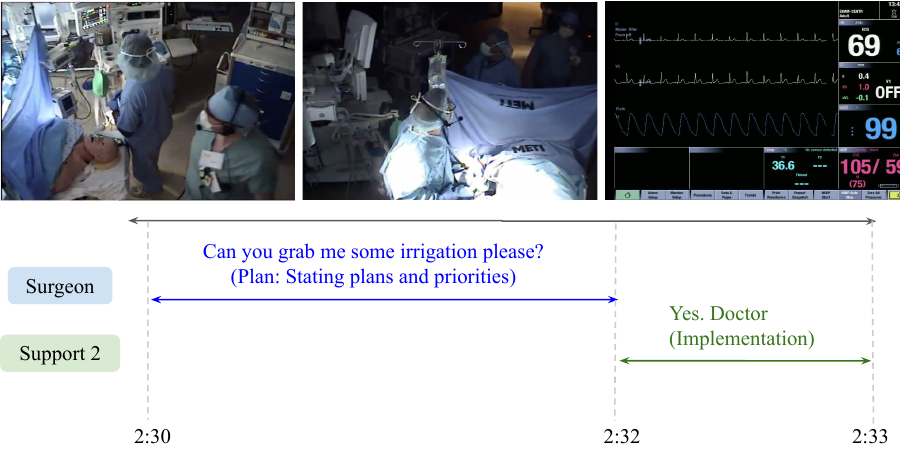}
        \caption{An example happens when the surgeon is conducting the operation, when the surgeon pauses and asks for support, and the support helps the surgeon to grab the irrigation.}
        \label{subfig: help-surgeon-example}
    \end{subfigure}%
    \caption{
    Additional examples of the labeled dialogue in the simulated operation.
    }
    \label{fig:more-examples}
\end{figure*}

\begin{figure}[th]
    \centering
     \begin{subfigure}[t]{0.45\textwidth}
        \centering
        \includegraphics[width=\linewidth]{figures/sunburst_analysis/sunburst-Support.crop.pdf}
        \caption{Support.}
    \end{subfigure}%
    \vspace{0.3em}
    \begin{subfigure}[t]{0.45\textwidth}
        \centering
        \includegraphics[width=\linewidth]{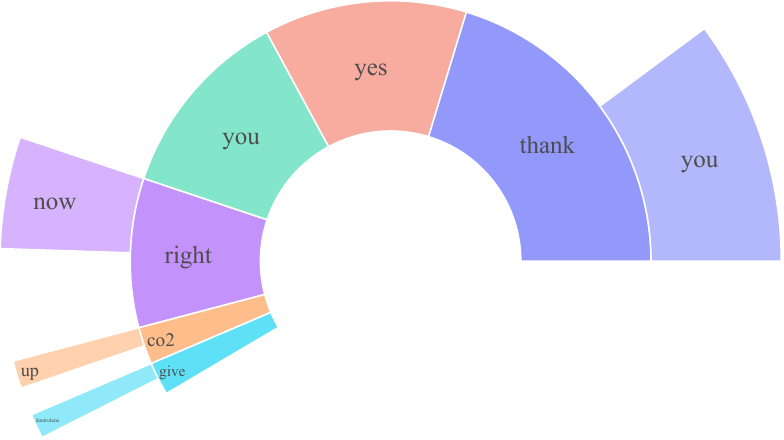}
        \caption{Trainee.}
    \end{subfigure}%
    \vspace{0.3em}
    \begin{subfigure}[t]{0.45\textwidth}
        \centering
        \includegraphics[width=\linewidth]{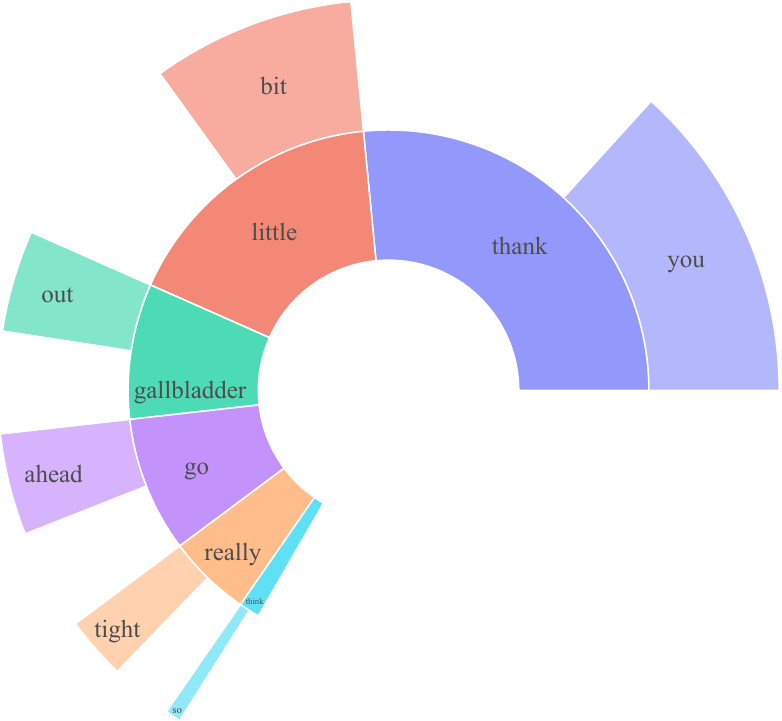}
        \caption{Surgeon.}
    \end{subfigure}%

    \caption{
    Distributions of words uttered from different roles in \dataname.
    }
    \label{fig:sunburst}
\end{figure}

\begin{figure}[th]
    \centering
     \begin{subfigure}[t]{0.45\textwidth}
        \centering
        \includegraphics[width=\linewidth]{figures/word_cloud_analysis/word_cloud-Support.pdf}
        \caption{Support.}
    \end{subfigure}%
    \vspace{0.3em}
    \begin{subfigure}[t]{0.45\textwidth}
        \centering
        \includegraphics[width=\linewidth]{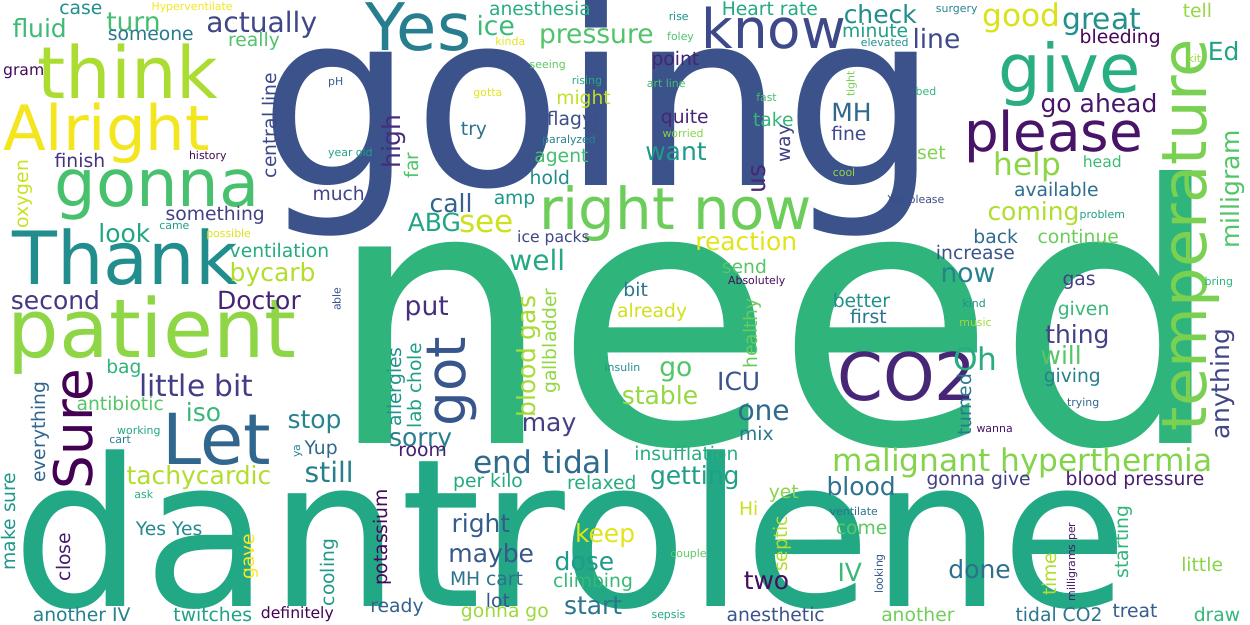}
        \caption{Trainee.}
    \end{subfigure}%
    \vspace{0.3em}
    \begin{subfigure}[t]{0.45\textwidth}
        \centering
        \includegraphics[width=\linewidth]{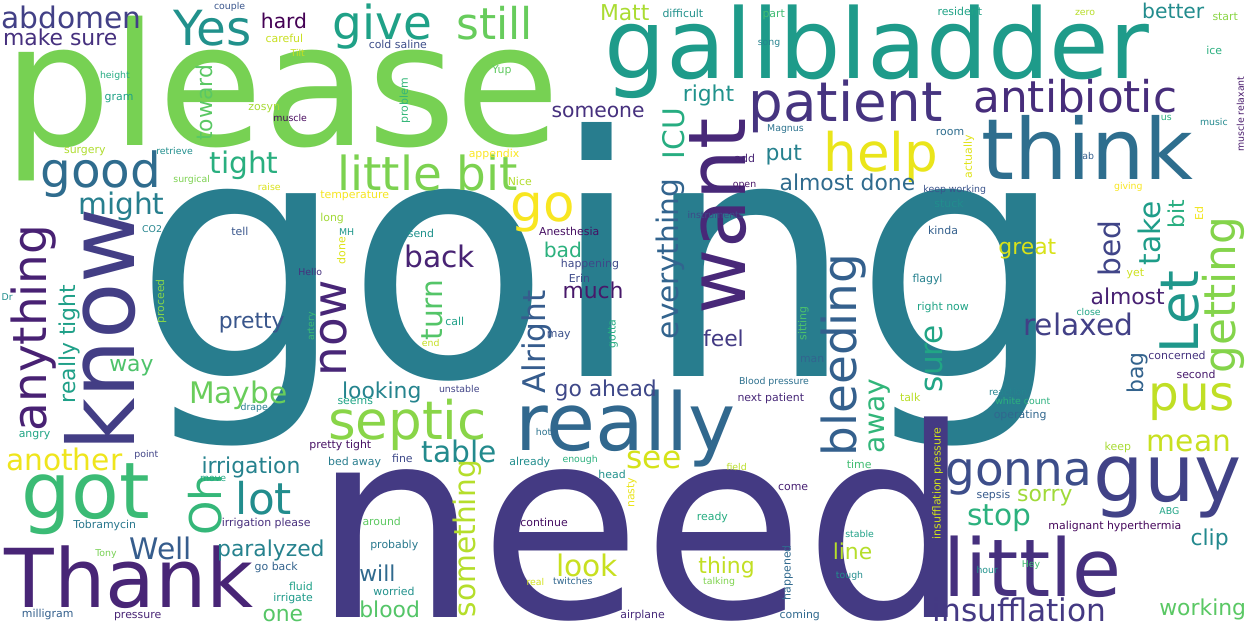}
        \caption{Surgeon.}
    \end{subfigure}%

    \caption{
    Word cloud plots for frequent words uttered by different roles involved in \dataname.
    }
    \label{fig:word-cloud}
\end{figure}

\Cref{fig:more-examples} provides two additional examples of the dialogues in our dataset.
\Cref{tab:dialogue-examples} provides dialogue snippets in our dataset.
\Cref{fig:sunburst} provides the word distributions for the three roles, respectively.
\Cref{fig:word-cloud} provides the most frequent words uttered by the three roles, respectively.

\paragraph{Specific word use.}
\Cref{tab:dantrolene-word-counts} provides the word counts for ``dantrolene'', we find that its frequency varies significantly. 
In one session, it appears 31 times, while in another, it occurs only once throughout the entire operation.
This variability suggests that certain medical terms naturally appear more frequently in specific cases rather than uniformly across all procedures.
Therefore, the presence of ``dantrolene'' does not indicate a dataset bias toward specific procedures but rather reflects real-world variations in clinical practice.

\begin{table}[t]
    \small
    \centering
    \begin{tabular}{cc}
    \toprule
    Session ID     & Counts \\
     \midrule
    1 & 8 \\
    2 & 12 \\
    3 & 10 \\
    4 & 10 \\
    5 & 8 \\
    6 & 31 \\
    7 & 9 \\
    8 & 18 \\
    9 & 14 \\
    10 & 16 \\
    11 & 3 \\
    12 & 4 \\
    13 & 13 \\
    14 & \textbf{31} \\
    15 & 20 \\
    16 & 17 \\
    17 & 13 \\
    18 & 3 \\
    19 & 6 \\
    20 & 19 \\
    21 & \textbf{1} \\
    22 & 12 \\
    \bottomrule
    \end{tabular}
    \caption{Word counts for ``dantrolene'' in the 22 sessions.
    We note that all the sessions last around the same.
    For instance, session 21 (``dantrolene'' appears once) lasts for 20:03 minutes, while session 14 (``dantrolene'' appears 31 times) lasts for 19:06 minutes.}
    \label{tab:dantrolene-word-counts}
\end{table}

\begin{table*}[t]
    \small
    \centering
    \begin{tabular}{cp{42em}}
        \toprule
        \textbf{Dialogue Snippet 1} \\
        \midrule
        ...\\
        Support 3 & So currently their recommendations are to not change the machine. Do you want to put the patient back on the vent and free up your hands? \\
        \grayc Trainee & \grayc Sure. Okay. If that's the recommendation. Absolutely. Great so next thing insulin, glucose, calcium.\\
        Support 2 & The ICU is calling.\\
        \grayc Trainee & \grayc Okay. Okay \\
        Support 1 & They're kinda just getting started. \\
        ...\\
        \midrule
        \textbf{Dialogue Snippet 2} \\
        \midrule
        ...\\
        Support 3 & Do you want to monitor her end-tidal with this?\\
        \grayc Trainee & \grayc That would be great? \\
        Support 3 & I gave 250 milligrams of the dantrolene.\\
        \grayc Trainee & \grayc Okay. Good. \\
        Support 2 & Here's the cold saline. I gotta go get the ice, okay? \\
        \grayc Surgeon & \grayc Thank you, Matt. \\
        Trainee & Dantrolene is in. We're going to cool the patient. The other thing is if we could. \\
        ... \\
        \midrule
        \textbf{Dialogue Snippet 3} \\
        \midrule
        ...\\
        Trainee & You've given how much neo? \\
        \grayc Support 1 & \grayc I've given probably- this really started maybe a few minutes ago- probably getting like 500mics. \\
        Trainee & Was she responding to it? \\
        \grayc Support 1 & \grayc She's responded a little bit. It's just kinda kept her around here but I think just because of the nausea, vomiting, and sepsis issue. \\
        Trainee & Okay. \\
        \grayc Support 1 & \grayc 36 year old lady who presented to the ER today with abdominal pain, nausea, vomiting. She was diagnosed with acute cholecystitis. They're afraid she's becoming septic. \\
        ...\\
        \bottomrule
    \end{tabular}
    \caption{
    Examples of the dialogues in \dataname.
}
    \label{tab:dialogue-examples}
\end{table*}


\section{More Details about the Methods}
\label{app-sec: more-model-details}

In addition to the methods in \Cref{sec: class-imblanace}, we have a majority vote baseline model which always predicts the major class. 
As expected, it reaches a decent micro F1 score (55.63) due to the class imbalance, while a much lower macro F1 score (14.01).
In addition, we test two non-deep learning methods such as RUSBoost \cite{seiffert2009rusboost} and SMOTE \cite{chawla2002smote} algorithm which is specifically designed to address class imbalance.
However, these pre-deep learning methods attains 24.21 and 32.32 macro F1 scores, much worse than simply tuning BERT$_\text{base}$ model or prompting LLMs.

\section{What Do Medical Professionals Expect from NLP?}
\label{app-sec: medical-expectation}



We are also interested to see how the medical professionals would view the results we get by employing these current NLP methods.
Therefore, we invite feedbacks from a medical professional who has been working in the domain for over a decade.
Here are what we get:
\begin{enumerate}[leftmargin=\parindent,align=left,labelwidth=\parindent,labelsep=0pt]
    \item They see a great opportunity to apply these LLMs on behavioral evaluation in the medical domain.
    They point out that the current evaluation practices in medical domains have significant limitations \cite{kolbe2019laborious, klonek2019time, stevenson2022development}, which typically are labor-intensive and prone to personal biases and errors. 
    They expect NLPers to develop consistent, reliable evaluation protocol to give feedback to the healthcare professionals.
    \item They expect a protocol that can take multimodal input into consideration including the team dialogue, patient vitals, and procedure videos.
    We note that this is one of the characteristics for \dataname.
    They also hope the NLP system could pinpoint specific teamwork deficiencies in the process.
    \item They also point out the related NLP methods that they find useful in their domain.
    For instance, intent classification, dialogue summarization, and multimodal reasoning works from NLP can provide quantifiable insights into teamwork dynamics and communication patterns in multimodal clinical data \cite{zhang2018team, allen2021insights, lehmann2023multimodal, hung2024discontent}.
    We note that \dataname~contain rich conversational data.
\end{enumerate}

\end{document}